\documentclass[letterpaper,11 pt]{article} 
\usepackage[margin=1in]{geometry}
\usepackage{times}
\usepackage{balance}
\usepackage[latin1]{inputenc}
\usepackage{amsmath}
\usepackage{amsfonts}
\usepackage{amssymb}
\usepackage{verbatim}
\usepackage{graphicx}
\usepackage{color}
\usepackage{hyperref}
\usepackage{subfigure}
\usepackage{bm}
\usepackage{url}
\usepackage{bbm}
\usepackage{balance}	
\usepackage{epsfig}
\usepackage{mathrsfs}
\usepackage{algorithm}
\usepackage{setspace}
\usepackage{algpseudocode}
\algnotext{EndFor}
\algnotext{EndIf}
\algnotext{EndWhile}

\newcommand{\mat}[1]{\boldsymbol{#1}}
\newcommand{\norm}[1]{\lVert#1\rVert_1}
\newcommand{\ER}{Erd\H{o}s-R\'{e}nyi }
\newcommand{\BA}{Barab\'{a}si-Albert }

\newcommand{\OG}{Ordinary-Greedy }
\newcommand{\LG}{Lazy-Greedy }
\newcommand{\SG}{Stochastic-Greedy }
\newcommand{\DG}{Distributed-Greedy }

\title{\LARGE \bf
Scaling Submodular Optimization Approaches for Control Applications in Networked Systems}

\author{Arun V Sathanur 
\thanks{Arun Sathanur is with the Physical and Computational Sciences Directorate,
        Pacific Northwest National Labs, Seattle, WA, USA
        {\tt\small arun.sathanur@pnnl.gov}}
}

\date{}

\begin{document}

\maketitle
\thispagestyle{empty}
\pagestyle{empty}


\begin{abstract}
Often times, in many design problems, there is a need to select a small set of informative or representative elements from a large ground set of entities in an optimal fashion. Submodular optimization that provides for a formal way to solve such problems, has recently received significant attention from the controls community where such subset selection problems are abound. However, scaling these approaches to large systems can be challenging because of the high computational complexity of the overall flow, in-part due to the high-complexity compute-oracles used to determine the objective function values. In this work, we explore a well-known paradigm, namely leader-selection in a multi-agent networked environment to illustrate strategies for scalable submodular optimization. We study the performance of the state-of-the-art stochastic and distributed greedy algorithms as well as explore techniques that accelerate the computation oracles within the optimization loop. We finally present results combining accelerated greedy algorithms with accelerated computation oracles and demonstrate significant speedups with little loss of optimality when compared to the baseline ordinary greedy algorithm. 
\end{abstract}

\section{Introduction}
Submodular optimization is increasingly becoming attractive to solve a number of controls problems involving subset selection with a number of theoretical advances in proving the submodularity of various types of objective functions \cite{summers2016submodularity, clark2017submodularity} encountered. Often times, the objective functions that involve computation of determinants or matrix inverses become extremely time-consuming for large problem sizes. The greedy algorithm has been commonly utilized for submodular maximization and it comes with theoretical approximation guarantees. However, in combination with expensive computation oracles, the problem sizes that can be tackled are still limited. Little work has been done in the area of scaling submodular optimization flows for larger problem sizes in controls applications. In this work, we focus on one particular widely-studied problem in controls namely the leader-selection in networked systems, make use of the recently proved submodularity results in this area and concentrate on the scalability aspects of the various flavors of the greedy algorithm used to solve submodular maximization. While we examine the scalability aspects from multiple angles, we also devote generous attention to the study of the loss of optimality of the solution from the baseline greedy algorithm when sampling is employed. 

Our application of focus is the selection of leader nodes in networked multi-agent systems. Scenarios with leader-follower dynamics span a wide variety of domains. Examples include unmanned autonomous vehicles \cite{notarstefano2011containment, liu2015leader}, gene regulatory networks \cite{liu2011controllability,lezon2006using} and social networks \cite{kempe2003maximizing, ghaderi2013opinion, parsegov2017novel}.  Such systems consist of interacting entities that can be represented in general by a weighted graph \cite{mesbahi2010graph}. A general problem of interest in many of the above domains is the ability to control the system by exercising external inputs to a subset of nodes called leader nodes. The idea is that the remaining non-input nodes or follower nodes are automatically guided to their desired states via local update rules (such as a weighted linear model). The input nodes could be selected based on application criteria such as noise performance or coherence-based metrics or consensus-related performance objectives \cite{fitch2016optimal, clark2017submodularity}.  

The optimization objective function that emerges in a number of scenarios is related to the graph Laplacian matrix \cite{fardad2011algorithms, clark2014supermodular, patterson2010leader, patterson2017optimal}. Different variations of the objective function exist, depending on the exact problem specification.  While  the work in \cite{fardad2011algorithms} considers a convex relaxation approach to solving the problem, references \cite{patterson2010leader} and \cite{clark2014supermodular} consider a greedy algorithm, whereas \cite{patterson2017optimal} develops graph algorithms for finding optimal solutions to specific topologies namely the path and ring graphs. In this work, we are concerned with an objective function that is given by the trace of the inverse of the grounded laplacian matrix, that is the sub-matrices of the Laplacian corresponding to the follower nodes \cite{clark2014supermodular, patterson2017optimal}. Our running example in this paper will use the supermodular minimization (equivalently submodular maximization since if $f$ is supermodular then $-f$ is submodular) approach for input node selection in order to minimize the overall noise variance of the system as outlined in \cite{clark2014supermodular}. 

\begin{figure}[htbp]
\includegraphics[width=8cm]{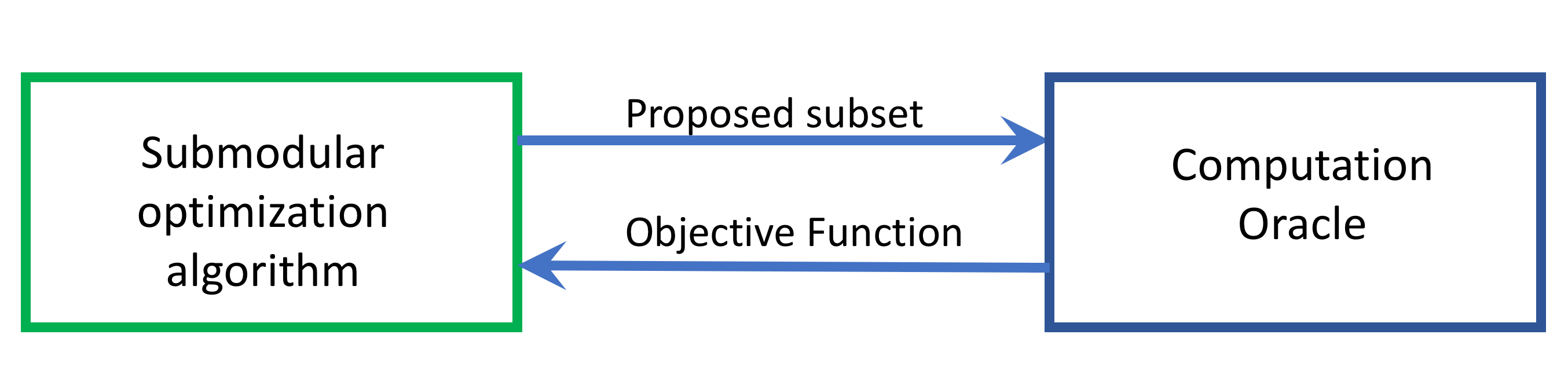}
\centering
\caption{A simplified diagram illustrating the interaction between a submodular optimization algorithm and the computation oracle that supplies the objective function. The overall complexity of the flow depends on the complexity of the optimizer and the oracle. System design flows benefit from scaling advances in both the pieces.}
\label{fig:fig1_submodular}
\end{figure}

Finding optimal subsets of a given size from a ground set, in order to maximize an objective function is in general NP-hard. Algorithms that come with approximation guarantees exist when the objective function is monotone submodular \cite{krause2014submodular}. The structure of such a submodular optimization flow in the most general case working with a black-box type objective function as computed by a computation oracle is shown in Fig. \ref{fig:fig1_submodular}.  
Our focus will be on the scalability of all aspects of the solution process, the quality of solutions obtained as well as exploiting the modular nature of the network topology when possible. The scalable approaches we consider are  are very general in nature and are applicable to a variety of sensing and control problems \cite{krause2008near, summers2016submodularity}. The algorithms that we will be considering are \OG\cite{nemhauser1978analysis}, \LG\cite{minoux1978accelerated, leskovec2007cost}, \SG \cite{mirzasoleiman2015lazier} and \DG \cite{mirzasoleiman2013distributed}. Of the four algorithms, \OG that has been previously leveraged in controls literature \cite{summers2016submodularity,clark2014supermodular} will be used as the baseline. Additionally, we will also be considering the effect of accelerating the computation oracle on the scaling performance of the above algorithms.

Section \ref{sec:prelims} describes the preliminaries and the optimization objective function for the leader selection problem. Section \ref{sec:submodular} introduces submodular functions along with the various flavors of the greedy algorithm examined in this work. Section \ref{sec:submod_results} discusses in detail, the performance of the various flavors of the greedy algorithm. Section \ref{sec:wsm} pairs the greedy algorithms with oracle acceleration and examines the resulting  scalability gains. Finally, Section \ref{sec:dist} considers the modular nature of networks and how this property can be utilized to leverage a two-stage distributed approach. We conclude this paper by summarizing the key findings and outline the future work in Section \ref{sec:conclu}.

\section{Preliminaries}
\label{sec:prelims}
Consider a networked multi-agent system $\bf \mathcal{M}$, represented as an undirected graph $G(V,E)$. The set of nodes denoted by $V$ represent the various agents that form the system. The set of edges $E$ refer to the connectivity between the agents. We denote the number of nodes by $n$ and the number of edges by $m$. The dynamics are in general represented as \cite{clark2014supermodular}
\begin{equation}
\dot{x_{i}}\left(t\right) = -\sum_{j \in N_i}\left( x_{i}\left(t\right) - x_{j}\left(t\right)\right) + w_{i}\left(t\right).
\end{equation}
Here $x_{i}\left(t\right)$ denotes the state of the node $i$ at time $t$ and $N_i$ denotes the neighborhood of the node $i$, i.e set of nodes $j$ such that $(i,j) \in E$. In general, we can also  incorporate weights $W_{ij}$ for the term $\left( x_{i}\left(t\right) - x_{j}\left(t\right)\right)$ in the above equation. $w_i\left(t\right)$ denotes the zero-mean unit variance Gaussian noise signal at each of the nodes $i$. Thus, in matrix form we can write: 
\begin{equation}
\boldsymbol{\dot{x}\left(t\right)} = \boldsymbol{-Lx}\left(t\right) +\boldsymbol{w}\left(t\right).
\end{equation}
Here $\boldsymbol{\dot{x}}\left(t\right)$ denotes the vector of time derivatives of the states which are denoted by the vector $\boldsymbol{x}\left(t\right)$. $\boldsymbol{L}$ denotes the graph laplacian that has the vertex degree values along it's diagonal and the negative of the weighted adjacency matrix forms the off-diagonal terms that is $\mat{L} = \mat{D} - \mat{A}$. Here $\mat{D}$ is the diagonal matrix of node degrees and $\mat{A}$ is the adjacency matrix of the graph.  $\boldsymbol{w}\left(t\right)$ is the vector of the noise signals described earlier. Let $\boldsymbol{l}$ be a binary vector of size $n$ that denotes whether a node is a leader ($1$) or not ($0$). Let $\boldsymbol{f}$ be the element-wise binary complement of the vector $\boldsymbol{l}$ that then indicates the follower nodes.  Given that the leader nodes are tied to external signals, we can then write the overall dynamics as:
\begin{equation}
\boldsymbol{\dot{x_f}}\left(t\right) = \boldsymbol{-L_{ff}x_f}\left(t\right) - \boldsymbol{L_{fl}x_l}\left(t\right) + \boldsymbol{w_f}\left(t\right).
\end{equation}
As demonstrated in \cite{fardad2011algorithms, clark2014supermodular, patterson2010leader, patterson2017optimal} the objective function for minimizing error metric  which is the deviation from the desired (consensus) state is the trace of the inverse of the grounded laplacian. The grounded laplacian is obtained by removal of specific rows and columns corresponding to the leader nodes  from the original laplacian matrix $\boldsymbol{L}$. Hence the optimization problem to choose $k$ leaders from the set of $n$ nodes, can be formulated as \cite{clark2014supermodular} follows. 
\begin{subequations}
\begin{alignat}{2}
&\!\min        &\qquad& \frac{1}{2}\boldsymbol{tr}\left(\boldsymbol{L}\left[\boldsymbol{f},\boldsymbol{f}\right]^{-1}\right)\\
&\text{subject to } &      & \norm{\boldsymbol{f}} = \left(n-k\right)
\end{alignat}
\end{subequations}
It is further shown \cite{clark2014supermodular} that the objective function is a monotone decreasing super-modular function implying that the minimization can be cast as maximizing $-f$ which is a monotone submodular function that allows us to explore the various flavors of greedy algorithms for the purpose.

\section{Submodular functions and Greedy algorithms}
\label{sec:submodular}
In this section we provide an overview of the algorithms that are used in this work. We refer the reader to the respective original references for all additional details. Consider a ground set of entities $V$ and let $S \subseteq V$. Let $f : 2^{V} \rightarrow \mathscr{R} $ be a mapping that associates every such $S$ with a real number. $f$ is submodular provided it satisfies the following. For every  $S \subseteq T \subseteq V$ and any $v \in V\setminus T$ we have $f(S \cup \{v\}) - f(S) \geq f(T \cup \{v\}) - f(T)$. Intuitively this property summarizes the notion of diminishing returns associated with the submodular functions. A second property that we are interested in the monotonicity property of a set function $f$ which is stated as follows. For every $S \subseteq T \subseteq V$, $f(S) \leq f(T)$. 

\subsection{Ordinary Greedy}
Consider the problem of maximizing a monotone submodular function under the cardinality constraint. That is $\max\limits_{S \subset V;  |S| \leq k} f(S)$. The algorithm to achieve this starts with an empty set for $S$ and proceeds by greedily selecting one element at each step such that the set of selected elements result in maximum marginal gain at each step. These steps are formalized in Algorithm \ref{algo:ord_greedy}. 
\begin{spacing}{1.8}
\begin{algorithm}[!htbp]
\begin{algorithmic}[1]
\Procedure{ORD-GREEDY}{$V,k,f : 2^{V} \rightarrow \mathscr{R}$}
\State $S$ $\gets$ $\{\}$, $i$ $=$ $0$
\While {$i$ $\leq$ $k$}
\State $m = 0.0$ 
\ForAll {$v$ $\in$ $V\setminus S$}
\If {$\left(f(S\cup{v}) - f(S)\right) > m$}
\State $s$ $\gets$ $v$ 
\State $m$ $=$ $\left(f(S\cup{v}) - f(S)\right)$
\EndIf
\EndFor
\State $S$ $\gets$ $S\cup s$
\EndWhile
\State \Return{$S$}
\EndProcedure
\end{algorithmic}
\caption{The \OG algorithm to maximize a monotone, submodular function with cardinality constraint.}
\label{algo:ord_greedy}
\end{algorithm}
\end{spacing}
It has been shown that the objective function value at the final step is at-least $\left(1-\frac{1}{e}\right)$ of the global optimal value for the same cardinality constraint \cite{nemhauser1978analysis}. Note that the computational complexity of the ordinary greedy algorithm is $\mathcal{O}(nk)$ calls to the computational oracle $f$. If the computational complexity of $f$ is $\mathcal{O}(n^d)$, the overall complexity of the optimization flow is given by $\mathcal{O}(n^{d+1}k)$. Making a reasonable assumption that $k$ grows with the size of the ground set (and is a small fraction of $n$), the overall complexity is therefore $\mathcal{O}(n^{d+2})$. Thus, for even moderate values of $d$ (1-2), the \OG algorithm is extremely limited in the size of the ground set, it can operate on. 

\subsection{Lazy Greedy}
The submodularity property means that the marginal gain of an element can never increase as the iterations proceed and that very often, the second best entity in the current iteration will be the best for the subsequent iteration. This means that, in the best case, if entity $u$ which is the second best in iteration $i$ has a marginal gain in iteration $i+1$ that is higher than the marginal gain of the third best entity in iteration $i$, then there is no need to  compute any of the additional marginal gains.  This observation can be used to avoid a large number of  computations of marginal gains thereby speeding up the greedy algorithm \cite{minoux1978accelerated,leskovec2007cost}. The \LG algorithm is detailed in Algorithm \ref{algo:lazy_greedy}. The data structures $\delta$ and $\lambda$ are priority queues that track the marginal gains and the status of the marginal gains (valid or invalid) for all the elements.  Note that the \LG algorithm is exact - it just avoids unnecessary function evaluations. In practice, the number of recomputations needed to pick the best element in subsequent iterations can vary from a few to a number close to the size of the ground set $n$. As such, the complexity of the \LG algorithm is indeterminate and is generally thought to be between $\left(\mathcal{O}(n)\right)$ and $\left(\mathcal{O}(nk)\right)$. The  \LG also becomes intractable for large problems, albeit a little later, compared to the greedy algorithm \cite{mirzasoleiman2015lazier}. 

\begin{algorithm}[!htbp]
\begin{algorithmic}[1]
\Procedure{LAZY-GREEDY}{$V,k,f : 2^{V} \rightarrow \mathscr{R}$}
\State $S$ $\gets$ $\{\}$, $i$ $=$ $0$
\ForAll {$s$ $\in$ $V$} $\delta\left(v\right)$ = $\infty$, $\lambda\left(v\right)$ = $False$
\EndFor
\While {$i$ $\leq$ $k$}
\State $m = 0.0$  
\ForAll {$v$ $\in$ $V\setminus S$} $\lambda\left(v\right)$ = $False$
\EndFor
\While {True}
\State $y$ $\gets$ $argmax_{s \in V\setminus S}$ $\delta\left(s\right)$
\If {$\lambda\left(y\right)$ = $True$}
\State $S$ $\gets$ $S\cup s$, $\delta$.pop($y$), $\lambda$.pop($y$)
\Else 
\State $\delta\left(y\right) \gets \left(f(S\cup{y}) - f(S)\right)$,$\lambda\left(y\right)$ = $True$
\EndIf
\EndWhile
\EndWhile
\State \Return{$S$}
\EndProcedure
\end{algorithmic}
\caption{The \LG algorithm to maximize a monotone, submodular function with cardinality constraint.}
\label{algo:lazy_greedy}
\end{algorithm}

\subsection{Stochastic Greedy}
\vspace{-0.6em}
Two recently proposed algorithms that significantly reduce the computational complexity of submodular optimization are MultGreedy \cite{wei2014fast} and \SG \cite{mirzasoleiman2015lazier}. \textit{MultGreedy} employ three different ingredients - an approximate version of  \LG, decomposing a given submodular into easy-to-evaluate surrogate submodular functions and finally, ground set pruning to filter out elements that can never be part of the optimal subset. \SG on the other hand subsamples the remainder of the ground-set at each iteration so that the overall complexity of the algorithm is reduced to $\mathcal{O}(n)$ calls to the computation oracle from $\mathcal{O}(nk)$ calls needed for the \OG algorithm. Due to it's simplicity and similar performance compared to the MultGreedy algorithm, we choose the \SG algorithm in our work. The main change from Algorithm \ref{algo:ord_greedy} is that before line $6$, the set $V\setminus S$ is subsampled without replacement by a factor given by $\beta = \frac{k}{log(1/\epsilon)}$. Here $0 < \epsilon < 1$ is the tradeoff parameter. Thus it provides for an reduction in the number of oracle calls by $\mathcal{O}(k)$ while providing a $\left(1-\frac{1}{e} - \epsilon\right)$ approximation guarantee in expectation \cite{mirzasoleiman2015lazier}. The full algorithm is detailed in Algorithm \ref{algo:stoch_greedy}. 
\begin{algorithm}[!htbp]
\begin{algorithmic}[1]
\Procedure{STOCH-GREEDY}{$V,k,\epsilon, f : 2^{V} \rightarrow \mathscr{R}$}
\State $S$ $\gets$ $\{\}$, $i$ $=$ $0$
\While {$i$ $\leq$ $k$}
\State $m = 0.0$, $\beta \gets \frac{k}{log(1/\epsilon)}$
\State $V_{s} \gets$ $sub\_sample(V\setminus S$,$\beta$)
\ForAll {$v$ $\in$ $V_{s}$}
\If {$\left(f(S\cup{v}) - f(S)\right) > m$}
\State $s$ $\gets$ $v$ 
\State $m$ $=$ $\left(f(S\cup{v}) - f(S)\right)$
\EndIf
\EndFor
\State $S$ $\gets$ $S\cup s$
\EndWhile
\State \Return{$S$}
\EndProcedure
\end{algorithmic}
\caption{The \SG algorithm to maximize a monotone, submodular function with cardinality constraint. The function $sub\_sample(A,\gamma)$ uniformly samples the given set $A$ without replacement so that the size of resultant sampled set is a fraction  $\frac{1}{\gamma}$ of the size of $A$.}
\label{algo:stoch_greedy}
\end{algorithm}
\vspace{-0.6em}

\subsection{Distributed Greedy}
A two-stage greedy approach was propose in \cite{mirzasoleiman2013distributed} for addressing scalability to very large datasets. The idea is to partition the dataset into $c$ partitions and run \OG on each of the $c$ partitions. This stage seeks $k$ representative elements from each of the partitions and collects the $ck$ elements as the ground set for a second round of greedy procedure. The second stage greedy algorithm down-selects the set of $ck$ elements to $k$ representative elements from the overall dataset. This algorithm called \textit{GreeDi} (for distributed greedy) can only provide a weakened approximation guarantee $\left(\frac{\left(1-\frac{1}{e}\right)^{2}}{min\left(m,k\right)}\right)$ which can be improved in certain cases involving decomposable functions (the function of the subset doesn't depend on the entire ground set) \cite{mirzasoleiman2013distributed}.  Thus one particular area of research is the empirical performance of this algorithm for problems such as the sensor placement and leader selection which in general cannot guarantee decomposable functions. We further note that in place of the \OG algorithm, in both the stages, we can use \LG and \SG algorithms as well with appropriate approximation guarantees.

In the next section, we will discuss various aspects of scaling related to the submodular optimization algorithms. This section uses a direct inverse computation routine for all the matrix inverse calculations. The following section (Section \ref{sec:wsm} considers the performance with the acceleration of the compute oracle. 

\section{Scaling performance of the submodular optimization algorithms}
\label{sec:submod_results}

In this section we demonstrate the speedups obtained applying the \LG and the \SG algorithms over the baseline algorithm namely, \OG. Note that \LG gives exactly the same results as the baseline but saves on extra computations. However, in the case of \SG, the deviation of the noise variance achieved by \SG as compared with \OG is also examined as a function of the tuning knob available in the form of the sampling fraction. 

The experimental setup consists of first generating a connected undirected graph following different topologies. For most of the experiments, we employ the \ER (ER) graph topology while we also investigate the \BA (BA) scale-free and the Random-Geometric (RG) models to illustrate specific points. In this section, all the matrix inverses are computed by treating the sparse Laplacian matrices as full and taking the direct inverse. Hence the objective function computation time doesn't depend on the structure of the sparse matrix involved. In the next section, we will specifically be considering the acceleration of the oracle computation. Each iteration of the algorithms will lead to one leader node being identified. The overall flow will proceed according the algorithm of choice (\OG,\LG or \SG). The computation of the objective function in each iteration $i+1$ involves the following steps. 
\begin{itemize}
\item Compute the grounded Laplacian $L_g$  by removing the rows and columns of the Laplacian corresponding to the set of leaders found till iteration $i$
\item Further remove the row and column corresponding to the candidate leader node for iteration $i+1$ (this step is different for each of the algorithms) to get the matrix of only the followers $L_{ff}$
\item Compute the objective function $tr\left(L_{ff}^{-1}\right)$
\end{itemize}

\begin{figure}[!htbp]
\begin{centering}
\includegraphics[width=0.9\columnwidth]{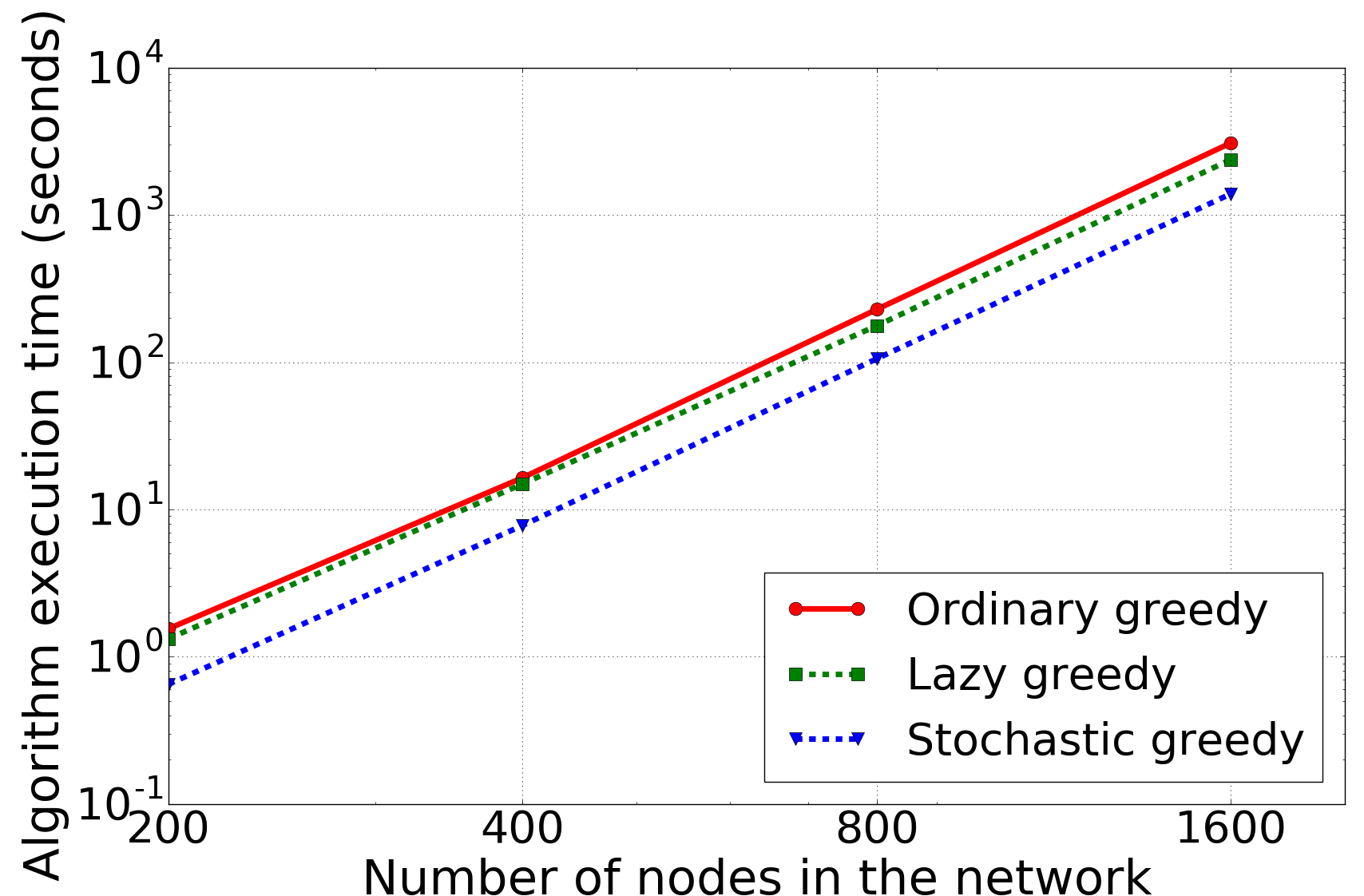}
\label{fig2_scaling_no_wsm}
\caption{Scaling performance of the three algorithms \OG,\LG and \SG for a fixed value of $k$.}
\vspace{-0.6em}
\end{centering}
\end{figure}

In our first experiment we vary the number of nodes in the (connected) network with an ER topology from 100 to 1600 and for each case, we seek $10$ leader nodes that minimize the noise variance.  For the \SG algorithm we assumed an $epsilon$ value of 0.01 to stay as close to the $(1-1/e)$ approximation guarantee. The execution times are depicted in Figure \ref{fig2_scaling_no_wsm} where both the axes are in logarithmic scale. 
Assuming that the run time of each of the algorithms is $an^dk$, where $a$ and $d$ are constants and $n$ being the problem size (number of nodes in the network), we can estimate the values of $a$ and $d$ via linear regression ($k$ is also a constant in this case). The slopes of each of the lines (corresponding to the exponent $d$) are approximately around 3.7 which is in line with the expected algorithm complexities. For example, the greedy algorithm requires $nk$ oracle calls. The oracle involves matrix inverse computation as the dominant cost and is approximately $\left(\mathcal{O}(n^{2.7})\right)$  (assuming $k << n$) \footnote{Our implementations of all the flows have been in Python using the numpy library for linear-algebraic operations. Careful experiments with profiling the matrix inversion routine returned a complexity of approximately $\left(\mathcal{O}(n^{2.7})\right)$}. In terms of actual speedups, the \LG algorithm achieves 1.2X and the \SG algorithm achieves 2.2X. The first inner-loop of the \SG algorithm uses sampling and for $k$=10 and $\epsilon$ = 0.01, the sampling factor $\beta$ = 2.17, which is exactly in-line with the speedup of 2.2 observed. Note that the second inner-loop involves the full matrix inversion and the complexity of that is unaltered. 

\begin{figure}[!htbp]
\begin{centering}
\includegraphics[width=0.9\columnwidth]{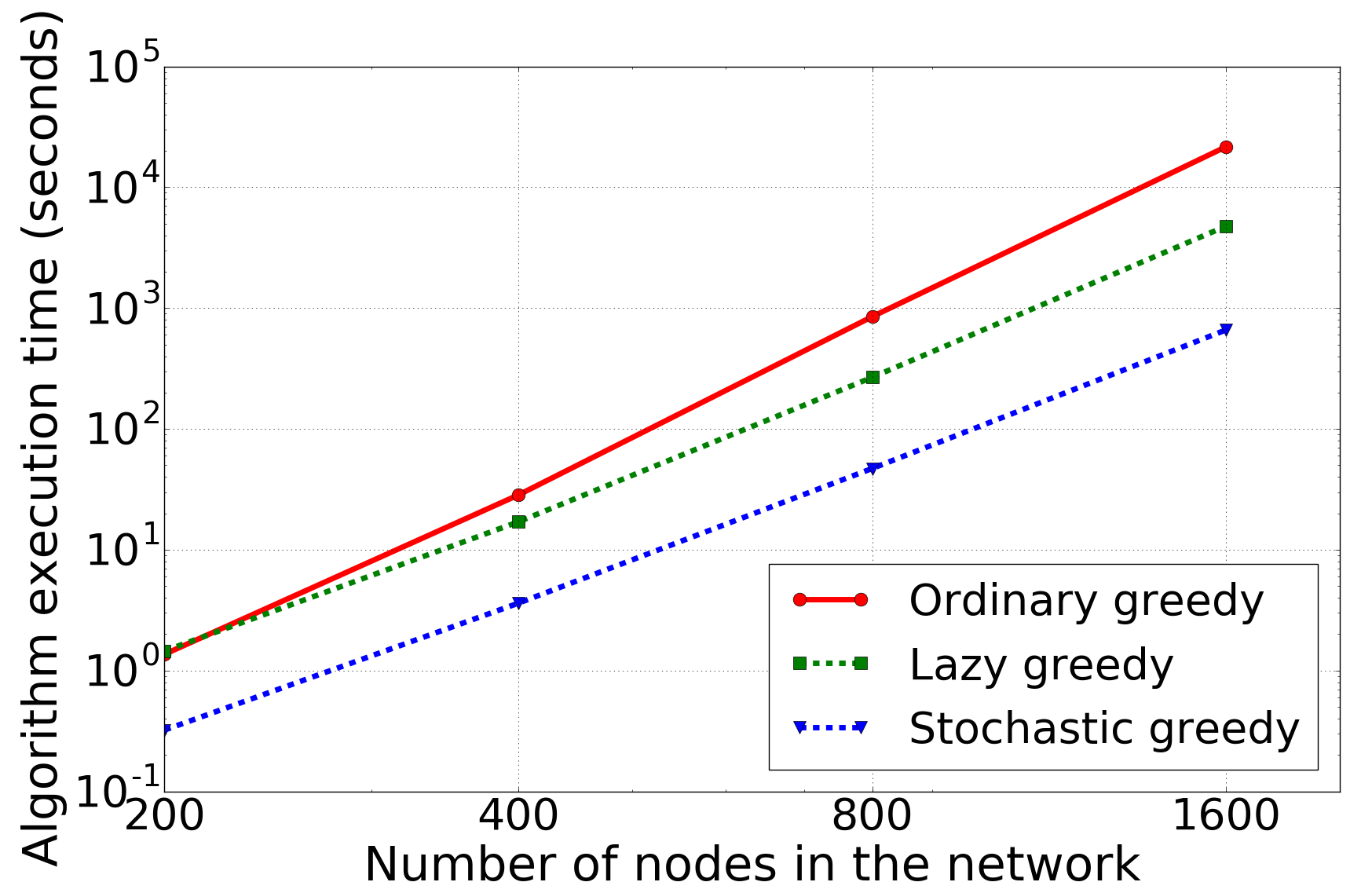}
\caption{Scaling performance of the three algorithms \OG,\LG and \SG when $k$=0.05$n$}
\label{fig3_scaling_no_wsm_prop}
\vspace{-0.6em}
\end{centering}
\end{figure}

Next we considered the growth of $k$ with $n$ instead of $k$ being a constant. This corresponds to a more realistic scenario where the number of leader nodes sought is a small fraction of the total number of nodes (i.e. it grows with the number of nodes). Setting the number of leader nodes to be 5\% of the total number of nodes, we achieved much better scalability with both \LG and \SG algorithms (for the ER topology and $\epsilon = 0.01$ for \SG). This is depicted in Figure \ref{fig3_scaling_no_wsm_prop} 
The \SG algorithm uses sampling for the second inner-loop and the sampling factor is proportional to the $k$ value thereby rendering the algorithm linear in $n$ for the number of oracle calls. Thus, the \SG algorithm is an order of magnitude faster ($d$ = 3.7) than the \OG algorithm ($d$ = 4.7) with maximum speedup of 32.8X obtained for $n$ = $1600$. The complexity of the \LG algorithm, in this scenario is between the \OG and \SG while it provides for a modest speedup of 4.5X for $n$ = $1600$.

\subsection{Performance of the \LG algorithm}
We do observe that the \LG algorithm does not save on too many computations in our example as opposed to other applications reported in literature such as in \cite{leskovec2007cost}. We explore the performance of the \LG algorithm by changing the topologies of the networks as well as increasing the $k$ value (number of leader nodes sought) since both of these can affect the number of recomputations. 
\begin{table}[h]
\caption{The speedup profile for the \LG algorithm.}
\begin{tabular}{||c||c|c|c|c|c||}
\hline
\hline
Graph & $k$ = 10 & $k$ = 20 & $k$ = 30 & $k$ = 40 & $k$ = 50 \\ [0.5ex] 
\hline
ER & 1.2X & 1.8X & 2.4X & 3.1X & 3.7X \\
RG & 1.8X & 3.0X & 4.0X & 5.0X & 5.8X \\
BA & 2.3X & 4.4X & 6.2X & 7.7X & 9.0X \\
\hline
\hline
\end{tabular}
\label{table1_models_k_speedup}
\end{table}
We summarize the analyses by tabulating the speedups relative to \OG for the various scenarios corresponding due graph models and $k$ values in Table \ref{table1_models_k_speedup}. Note that the total number of nodes in the networks were fixed at 500 for all the cases.  Larger $k$ values and a skewed degree distribution for the BA graph (power-law)  as opposed to a rather homogeneous degree distribution (Poisson) for the ER graph results in fewer recomputations and a better speedup for the \LG algorithm.

\subsection{Speed-Optimality tradeoff for the \SG algorithm}
\begin{figure}[!htbp]
\begin{centering}
\includegraphics[width=0.9\columnwidth]{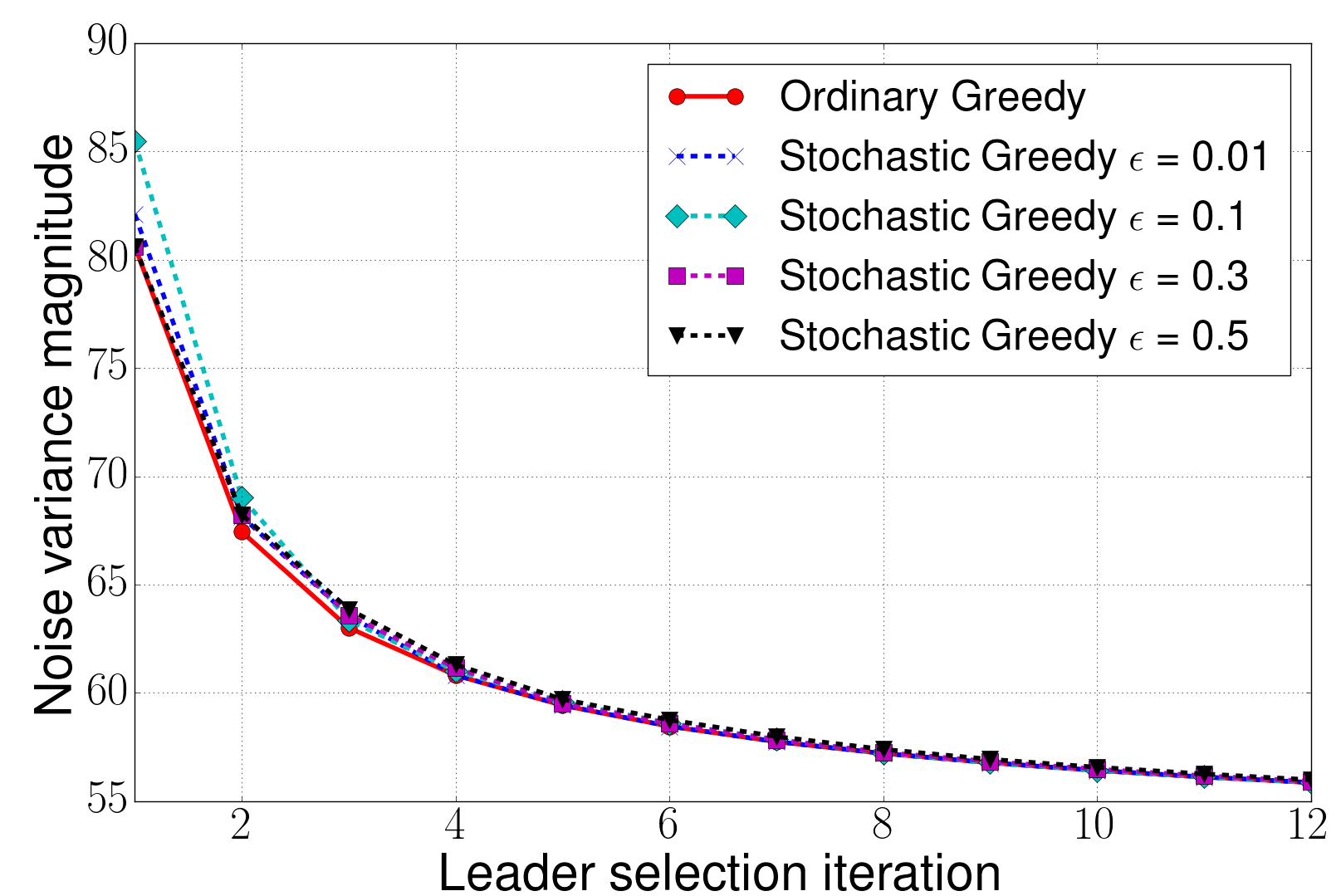}
\caption{Noise optimality achieved by the \SG algorithm for various values of $\epsilon$ for $n$=1000 and $k$ =12}
\label{fig4_nfigure_sg}
\vspace{-0.6em}
\end{centering}
\end{figure}

While the \LG algorithm is exact (same solution as the \OG algorithm) and saves on additional computations, the \SG algorithm, on the other hand, is able to trade-off the sampling factor with the optimality through the use of a constant $\epsilon$. Thus we expect that, as we increase $\epsilon$, the optimal solution obtained by the \SG algorithm gets progressively worse compared to the \OG algorithm while the execution time goes down linearly with $\log\left(\frac{1}{\epsilon}\right)$. In practice, the decrease in optimality is pronounced only for the first few leader nodes selected. Our next experiment confirms this observation and it shows that even for values of $\epsilon$ as large as 0.5, and for modest $k$ values such as 10, the noise variance degradation is nearly absent where as we obtain excellent speed-ups as far as the execution time is concerned. Figure \ref{fig4_nfigure_sg} shows the noise variance achieved for the number of leaders up to 12 ($k$ = 12) for a $n$ = 1000 node network. We can see that except for the first few leader nodes, the noise optimality achieved by the \SG algorithm is virtually indistinguishable from that of the \OG algorithm. 
The speedups achieved in the four cases corresponding to $\epsilon = 0.01$, $\epsilon = 0.1$, $\epsilon = 0.3$ and $\epsilon = 0.5$ are 2.7X,  5.3X, 7.6X and 17.7X respectively. For the BA graph, we noticed a similar behavior except that it look a larger value of $k$ before the noise variance values achieved by the \SG algorithm approached to less than 1\% of the value achieved by \OG. 

\begin{figure}[!htbp]
\begin{centering}
\includegraphics[width=0.9\columnwidth]{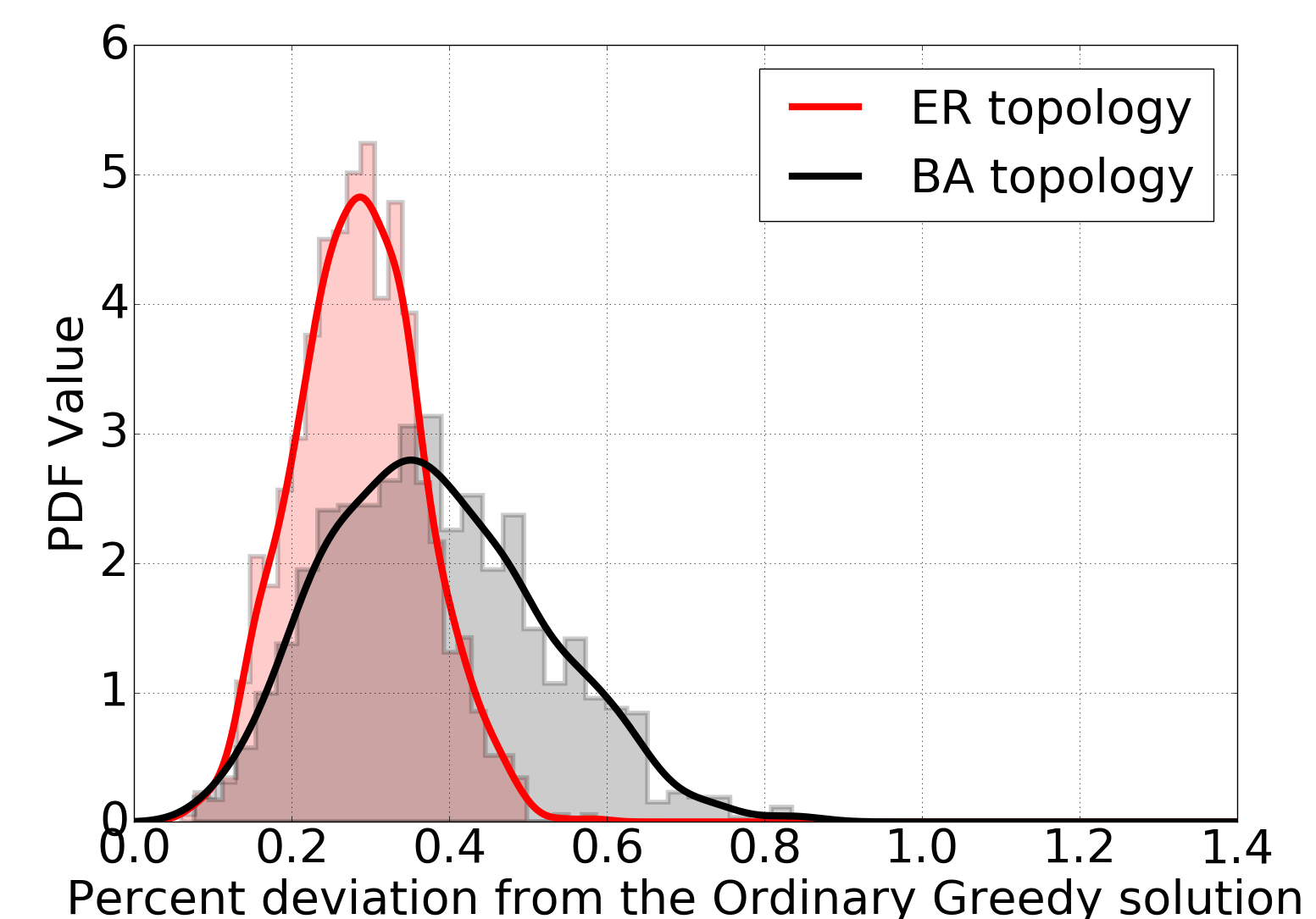}
\caption{Distributions of the percentage deviations of the performance objective achieved by \SG with respect to that of \OG at $k$ = 12 for 1000 node graphs (ER and BA). We ran 1000 Monte Carlo trials. }
\label{fig6_noise_variance}
\vspace{-0.6em}
\end{centering}
\end{figure}

An important step in the \SG algorithm is the sampling step in the inner loop of the algorithm which gives it's speedup. This step involves random sampling without replacement and hence is dependent on the random seed. The previous observation that the degradation of the performance objective is minimal is made based on a single run (one particular random seed value). To ensure that this is not a one-off observation, for the same $\epsilon$ = $0.5$, we ran a Monte Carlo experiment where we ran 1000 experiments with different random seeds and measured the percentage deviation of the solution obtained by the \SG algorithm from that of the \OG algorithm at $k$ = 12 for both an \ER graph and a \BA graph of 1000 nodes each. Figure \ref{fig6_noise_variance} shows the histograms (and smoothed densities) of the deviation and therefore verifies that the performance degradation of the \SG algorithm is minimal.

In summary, we demonstrated the performance of three different types of greedy algorithms for the leader-selection problem and noted that, in \SG provides an order of magnitude improvement in the execution time (for a typical scenario we also demonstrated a 32.8X speedup) over the \OG algorithm while producing solutions with very little degradation in the performance objective. The \LG algorithm on the other hand was shown to provide reasonable speedups for larger values of $k$ and it achieves these speedups while providing the exact same results as \OG. 

\section{Accelerating the oracle computation}
\label{sec:wsm}

In Section \ref{sec:submod_results} we demonstrated the performance of various types of greedy algorithms for the leader-selection problem. Even though we demonstrated 32.8X speedup for the \SG algorithm over \OG, the overall complexity of the algorithm is still high and the main contribution to this is the complexity of the oracle which involves matrix inversion. This section demonstrates how several known techniques can be combined to reduce the complexity of the oracle so as to obtain additional speedups. Note that the acceleration of the oracle still provides us with the \textit{exact inverse}. Hence all the solutions obtained via \OG and \LG algorithms remain the same but they are executed much faster. The solutions obtained via the \SG algorithm will also remain the same (assuming same random seeds) and it shows the same behavior with respect to optimality as compared to the \SG algorithm without the oracle acceleration.

At any general iteration $\left(i+1\right)$ where we select the $\left(i+1\right)^{th}$ leader node, and for each of the three greedy algorithm variants that we have been studying, we need to repeatedly compute the inverse of the matrix $L^{i+1}_g$ that is obtained by removing exactly one row and one column (corresponding to the candidates) from the optimal (till iteration $i$) $L^{i}_g$ matrix found in iteration $i$.  Rather than computing the inverse directly, each time, we can update the inverse found in the iteration $i$ by making use of the Woodbury matrix identity \cite{hager1989updating}. Also, instead of the usual low-rank update to the matrix, we will be leveraging the identity to compute the inverse of a matrix with a row and column removed (same indices for the row and column in this application), i.e. the inverse of the matrix $L^{i+1}_g$ from the inverse of the matrix $L^{i}_g$. One caveat is that, in the case of graph Laplacians, this process is applicable for all iterations other than the first iteration since the Laplacian doesn't have an inverse where as all grounded Laplacians (Laplacian matrix with rows and columns corresponding to desired nodes removed) have unique inverses \cite{pirani2014spectral}. 

Consider an invertible square matrix $\mat{A}$ whose inverse $\mat{A}^{-1}$ is already computed. We remove one row and column with the same index (say the $m^{th}$ row and column) of this matrix resulting in the matrix $\mat{A}_{m}$ and we wish to update the matrix $\mat{A}^{-1}$ to obtain $\mat{A}_{m}^{-1}$. We first interchange the $m^{th}$ row and column of $\mat{A}$ with the last row and column to obtain $\mat{A}^{'}$. The inverse of $\mat{A}^{'}$ will be $\mat{A}^{-1}$ with it's $m^{th}$ row and column interchanged with the last row and column in a similar manner. Now we update the matrix $\mat{A}^{'}$ with a rank-2 update $\mat{U}\mat{V}^T$, to get the matrix $\mat{B}$ so that the last row and columns of $\mat{B}$ are zero except for the last element $a_{ll} = \mat{A}^{'}\left(l,l\right)$ where $l$ is the size of $\mat{A}$ (or $\mat{A}^{'}$). Note that $\mat{U}$ and $\mat{V}$ are of dimensions $l \times 2$ and $2 \times l$ and the entries are appropriately chosen to yield the desired form of $\mat{B}$. 

\begin{equation}
\mat{B} =  \begin{bmatrix} \mat{A}_{m} & \mat{0} \\ \mat{0} & a_{ll}  \end{bmatrix} = \mat{A}^{'} + \mat{U}\mat{V}^T
\end{equation}
\begin{multline}
\mat{B}^{-1} =  \begin{bmatrix} \mat{A}^{-1}_{m} & \mat{0} \\ \mat{0} & \frac{1}{a_{ll}}  \end{bmatrix} \\
		   = \mat{A}^{'-1} - \mat{A}^{'-1}\mat{U}\left(\mat{I}+\mat{V}\mat{A}^{'-1}\mat{U}\right)^{-1}\mat{V}\mat{A}^{'-1}
\label{eqn_woodbury}
\end{multline}
Note that in the above equation, the inverse of the low-rank update to $\mat{A}^{'}$ is expanded out via the Woodbury matrix identity\cite{hager1989updating}. 
The top $(l-1) \times (l-1)$ sub-matrix of $\mat{B}^{-1}$ is the desired $\mat{A}_{m}^{-1}$. The computational complexity of the most expensive sub-task in the Woodbury identity is the multiplication of matrices of size $l \times 2$ and $2 \times l$ resulting in $2l^{2}$ multiplications. For the case of our grounded Laplacians, assuming $k <<  n$, we see that the updating of the inverse from iteration $i$ to $i+1$ (for $i < k$) will result in approximately $2n^2$ computations. 

We now tackle the question of updating the inverse for the very first iteration (selection of the first leader node). The low-rank update that we outlined above, will not be applicable in this scenario since the graph Laplacian does not have an inverse (it's first eigenvalue is always $0$). We leverage the strategy developed in \cite{xiao2003resistance,ranjan2014incremental} that makes use of the properties of the graph Laplacian in finding the inverse of the first grounded Laplacian from the Moore-Penrose psuedo-inverse of the Laplacian $\mat{L}^{\dagger}$. 

\begin{figure}[!htbp]
\begin{centering}
\includegraphics[width=0.9\columnwidth]{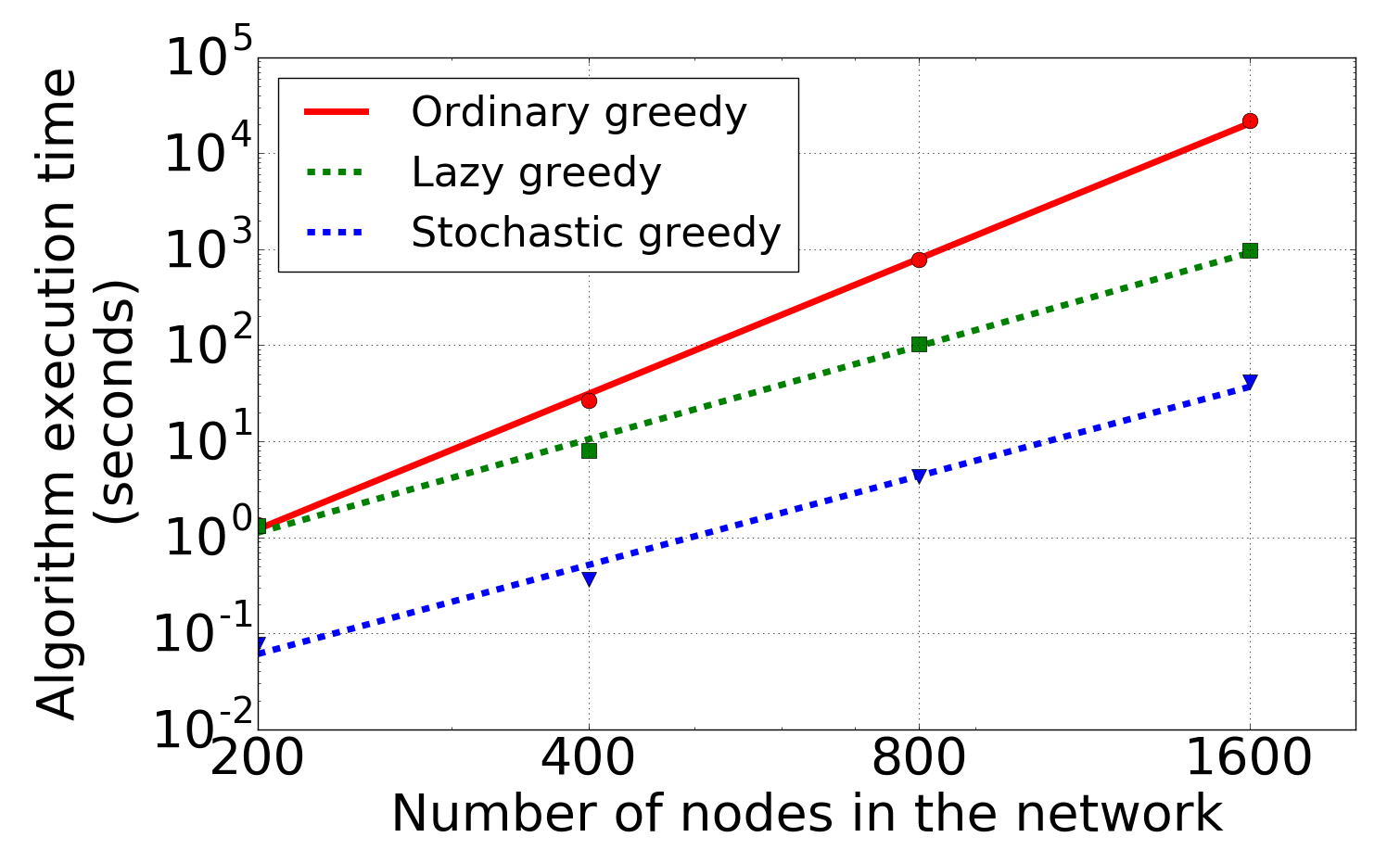}
\caption{Scaling performance of the three algorithms \OG,\LG and \SG when $k$=0.05$n$ and when oracle acceleration is applied to \LG and \SG}
\label{fig5_scaling_wsm_prop}
\vspace{-0.6em}
\end{centering}
\end{figure}

Consider the $n \times n$ graph Laplacian matrix $\mat{L}$ as described in Section \ref{sec:prelims}. It can be shown \cite{ranjan2014incremental} that the pseudo-inverse of $\mat{L}$, $\mat{L}^{\dagger}$ can be found by a very simple procedure as follows.
\begin{equation}
\mat{L}^{\dagger} = \left(\mat{L} + \frac{1}{n}\mat{J}\right)^{-1} - \frac{1}{n}\mat{J}
\label{eqn_pinv_new}
\end{equation}
where $\mat{J}$ is an $n \times n$ matrix of all $1^{s}$. Note that, while the complexity of computing the psuedo-inverse is $\mathcal{O}(n^{3})$  via both Equation \ref{eqn_pinv_new} as well as via the Singular Value Decomposition (SVD) based methods, in practice the execution time for arbitrary graphs is much smaller for Equation \ref{eqn_pinv_new} as demonstrated in \cite{ranjan2014incremental}. Once $\mat{L}^{\dagger}$ is found, the inverse of the first grounded Laplacian obtained by removing the $n^{th}$ row and column from $\mat{L}$ is given by the following procedure \cite{ranjan2014incremental}.
\begin{equation}
\left[\mat{L}_{n}^{-1}\right]_{xy} = \mat{L}^{\dagger}_{xy} - \mat{L}^{\dagger}_{xn} - \mat{L}^{\dagger}_{ny} + \mat{L}^{\dagger}_{nn}
\label{eqn_Lg}
\end{equation}
For the removal of an arbitrary $m^{th}$ row and column of $\mat{L}$ ( $m < n$), the inverse can be found via interchanging the row and column in question with the 
$n^{th}$ row and column  respectively and re-applying Equation \ref{eqn_Lg}. All the inverses required by the first iteration are therefore computed in $\mathcal{O}(n^{2})$ time, each. Thus the complexity of the \SG algorithm with this acceleration of the oracle becomes $\mathcal{O}(n^{3})$ as opposed to $\mathcal{O}(n^{4}k)$ for the \OG algorithm and if $k$ is a fraction of $n$ as discussed earlier, the \SG algorithm with the oracle acceleration delivers two orders of magnitude speedups while resulting in very little loss of optimality of the solution as compared \OG. These results are illustrated in Fig. \ref{fig5_scaling_wsm_prop}. 
In terms of raw speedups, \SG with oracle acceleration and $\epsilon$ = $0.5$ provides a 517X speedup over \OG for a network with $1600$ nodes and seeking $80$ leader nodes while the deviation from the \OG solution is less than 1\% for $k > 10$ leader nodes. A quick analysis shows a speedup of the order of $\mathcal{O}(n^{1.7})$ and this is confirmed by regression analysis on the wall-times for each of the algorithms which shows a speedup order of magnitude in the range of $\mathcal{O}(n^{1.6})$ for \SG over \OG. 

\section{Distributed Submodular Optimization}
\label{sec:dist}
For very large networks ($n$ $>$ 10,000 say), that occur in certain applications such as social communities, even the \SG algorithm becomes intractable. For example, extrapolating the regression line in Figure \ref{fig5_scaling_wsm_prop}, when the network size is 20000, the algorithm running time is noted to be about 1 day where as for a network of size 40,000 the running time becomes more than a week. In this case we can make use of the \DG \cite{mirzasoleiman2013distributed} algorithm where we apply the greedy algorithms in two stages. Assuming that the graph can be decomposed into $c$ clusters, the first stage will be used to accumulate $k$ candidate leader nodes from each of the clusters into a single list of $ck$ candidates. The second stage will be used to refine the selection to $k$ leader nodes from the list of $ck$ candidates. Note that we can use any combination of the greedy algorithms that we discussed above with or without the oracle acceleration for the two stages of \DG. 

\begin{figure}[!htbp]
\begin{centering}
\includegraphics[width=0.9\columnwidth]{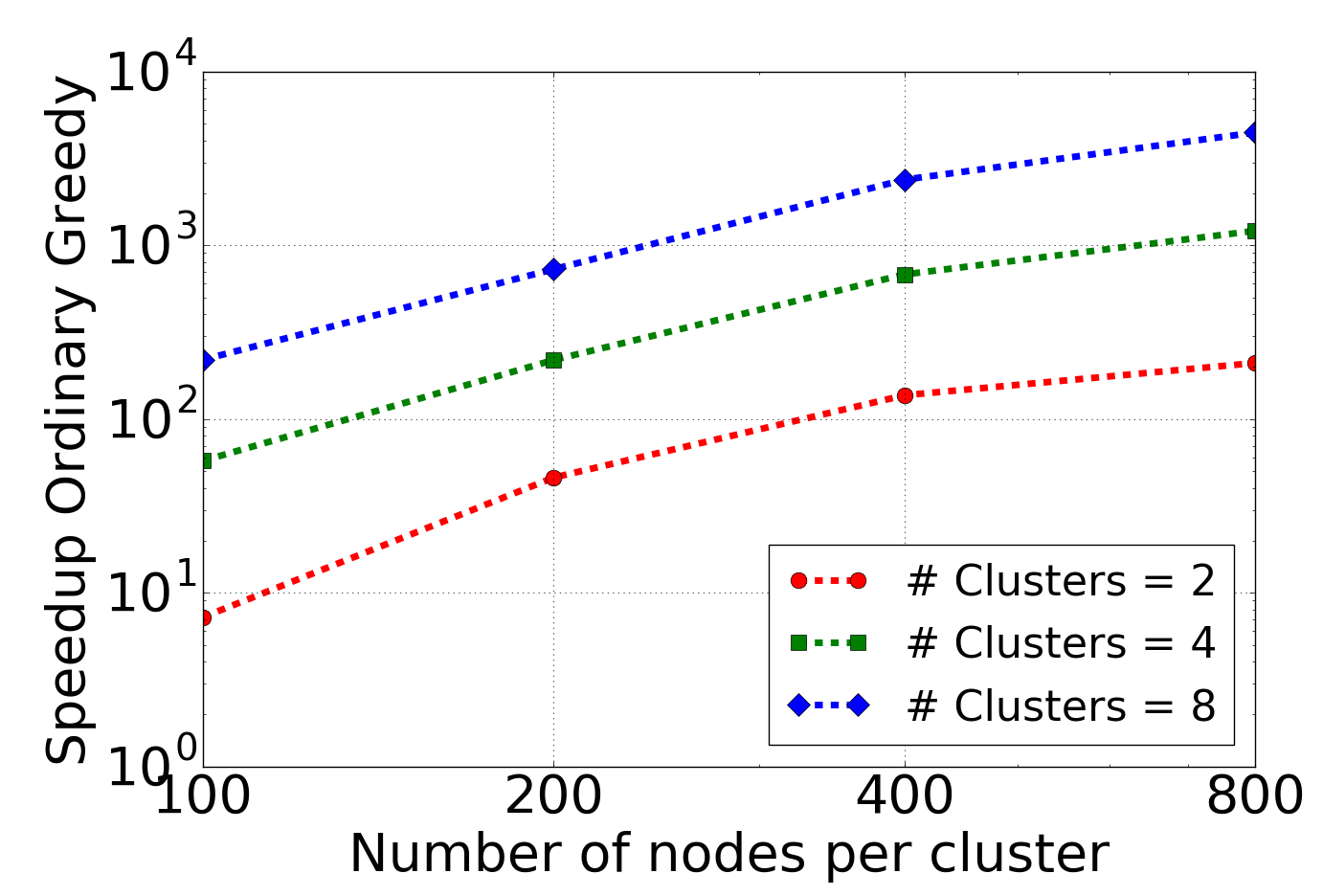}
\caption{Scaling performance of the \DG algorithm for different cluster sizes and number of clusters. \DG was implemented with \SG and the low-rank acceleration.}
\label{fig7_cluster_scaling}
\vspace{-0.6em}
\end{centering}
\end{figure}

We first generated a graph with dense clusters and sparse inter-cluster edges based on the stochastic block model \cite{mossel2012stochastic}. The number of nodes per cluster was $n_C$ and there were $c$ equal-sized clusters in the graph. We also set the edge probability to be $0.05$ for intra-cluster connectivity ($p_{in}$) and $0.02$ for inter-cluster connectivity ($p_{out}$). The number of leader nodes sought ($k$) was 10.  For the \DG implementation we utilized the \SG algorithm with $\epsilon$ = $0.5$ for both the stages.  The baseline for each configuration is \OG algorithm running on the entire stochastic block model graph with ($cn_C$ nodes).  Figure \ref{fig7_cluster_scaling} shows the speedups as a function of $n_C$ for different values of $c$. It's clear that both the number of clusters and the cluster size have an effect on the speedup and the speedups approach 5000X+ for 800 nodes with 8 clusters. Note that for the overall graph sizes greater than 1600 nodes, for the \OG algorithm, we used the predicted value of the running time from the regression model introduced in Section \ref{sec:submod_results}. The exact analysis of the algorithm complexity is out of scope of this work.

Given the utilization of of the  \SG algorithm in both the stages, a natural question that arises is how bad is the deviation of the noise variance from the solution obtained by the \OG algorithm (or it's exact equivalents). Surprisingly, we did not find any evidence of even moderate degradation in the performance beyond the first few leaders. As demonstrated in Section \ref{sec:submod_results}, the deviation from the \OG solution for the full graph was large at very small values of $k$ but for a reasonably small value of $k$, namely, $k$ = $10$, the deviation was less than 1\%. Another interesting experiment we performed was to change the value of $p_{out}$ from 0.2$p_{in}$ to 1.0$p_{in}$ and  we observed the the deviation from the \OG solution for the full graph at $k$ = $10$ for all the cases. This has the effect of increasing the inter-cluster edges so that the clusters lose their identity and the modular graph becomes progressively homogeneous. In all the cases, we observed that the deviation at $k$ = $10$ was less than 1\%.  This implies that, even for an arbitrary graph, without modular structure, there is a possibility that mere partitioning of the graph into equal-sized partitions might be sufficient to take full advantage of the \DG algorithm. This will circumvent the issue of suboptimal scaling when the cluster sizes are non-uniform or even have a highly-skewed distribution. A detailed investigation in this direction is beyond the scope of this paper. 

\section{Conclusions and Future Work}
\label{sec:conclu}
In this work, using the leader-selection problem as an example, we examined the scaling and optimality aspects of various flavors of greedy algorithms used to solve submodular maximization. We investigated in detail, the performance of the \LG algorithm which is exact and the \SG algorithm which uses a sampling step to provide scalability. We demonstrated that the performance of \SG can be significantly faster compared to \OG and \LG and this comes at the cost of only a small loss in optimality. We then noted that, in conjunction with the low-rank updates for computing the inverses of the grounded Laplacian matrices, \LG and \SG can provide even more speedups. Finally, we examined the performance of the \DG algorithm which involves a two-stage greedy strategy. In addition to demonstrating very good scalability, a detailed investigation shows minimal loss of optimality with respect to \OG. In the process we also made a very interesting observation on leveraging graph-partitioning to apply the \DG strategy to arbitrary graph topologies that lack modular structure. 

Future work is focussed on a more diverse set of objective functions stemming from different applications, more sophisticated update methods, parallelization of the implementation for high performance computing architectures and exploration of graph partitioning for optimal scaling with multi-stage greedy methods.

\end{document}